\pdfoutput=1
\documentclass[11pt]{article}

\usepackage[dvipsnames,table]{xcolor}
\usepackage[preprint]{acl}

\usepackage{times}
\usepackage{latexsym}

\usepackage[T1]{fontenc}
\usepackage[utf8]{inputenc}

\usepackage{microtype}

\usepackage{inconsolata}

\usepackage{graphicx}

\usepackage{hyperref}       % hyperlinks
\usepackage{url}            % simple URL typesetting
\usepackage{booktabs}       % professional-quality tables
\usepackage{amsfonts}       % blackboard math symbols
\usepackage{nicefrac}       % compact symbols for 1/2, etc.
\usepackage{microtype}      % microtypography
\usepackage{amsmath}
\usepackage{amssymb}
\usepackage{mathtools}
\usepackage{amsthm}
\usepackage{bbm}
\usepackage{thmtools}
\usepackage{tabularx}
\usepackage{cleveref}

\usepackage{multirow}
\usepackage{caption}

\theoremstyle{plain}
\newtheorem{theorem}{Theorem}[section]

\theoremstyle{definition}
\newtheorem{definition}[theorem]{Definition}

\theoremstyle{remark}

\newcommand\numberthis{\addtocounter{equation}{1}\tag{\theequation}}

\newcommand{\ours}{thought calibration}
\newcommand{\ourscaps}{Thought calibration}

\usepackage{mdframed}
\mdfdefinestyle{prompt}{%
linecolor=white,
outerlinewidth=0,
innertopmargin=4pt,
innerbottommargin=4pt,
innerrightmargin=4pt,
innerleftmargin=4pt,
leftmargin = 0pt,
rightmargin = 0pt,
backgroundcolor=black!5,
}

\newenvironment{prompt}{
\setlength{\parskip}{2pt}
\begin{mdframed}[style=prompt]
\normalfont\fontfamily{cmvtt}\selectfont\scriptsize}{\end{mdframed}\par}

\newcommand{\fstring}[1]{\textcolor{blue}{\{#1\}}}

\title{\ourscaps{}:\\
Efficient and confident test-time scaling}

\author{Menghua Wu, Cai Zhou, Stephen Bates, Tommi Jaakkola \\
Department of Computer Science\\
Massachusetts Institute of Technology\\
Cambridge, MA 02319 \\
}

\begin{document}

\maketitle

\begin{abstract}

Reasoning large language models achieve impressive test-time scaling by thinking for longer, but this performance gain comes at significant compute cost.
Directly limiting test-time budget hurts overall performance, but not all problems are equally difficult.
We propose \emph{\ours{}} to decide dynamically when thinking can be terminated.
To calibrate our decision rule, we view a language model's growing body of thoughts as a nested sequence of reasoning trees, where the goal is to identify the point at which novel reasoning plateaus.
We realize this framework through lightweight probes that operate on top of the language model's hidden representations, which are informative of both the reasoning structure and overall consistency of response.
Based on three reasoning language models and four datasets,
\ours{} preserves model performance with up to a 60\% reduction in thinking tokens on in-distribution data, and up to 20\% in out-of-distribution data.
\end{abstract}

\section{Introduction}
\label{section:intro}

Test-time scaling presents a new paradigm for improving language model reasoning by expending large amounts of compute during inference~\citep{kaplan2020scaling,wei2022chain}.
Though the strategies for eliciting reasoning vary -- from large-scale reinforcement learning~\citep{guo2025deepseek} to explicit tree search~\citep{zhang2024accessing,zhang2024rest} -- a common effect is that language models improve by sampling substantially more tokens.
This may result in wasted compute on easy problems~\citep{chen2024not,sui2025stop}, but naively limiting the generation length leads to pronounced drops in accuracy~\citep{muennighoff2025s1}.
This motivates early stopping strategies that reduce the inference budget without significantly degrading performance, and control the extent of impact, if performance must be compromised.

Numerous methods have been proposed for teaching language models to be economical with their token budgets~\citep{han2024token,arora2025training,sui2025stop}, or for identifying opportune stopping points~\citep{yang2025dynamic,zhang2025reasoning}.
While these methods demonstrate strong empirical performance, they lack strict statistical guarantees about when they could fail.
In an orthogonal direction, conformal prediction has been adapted to equip language models with calibrated confidences about the quality or consistency of their generations~\citep{tatsu2024factuality,quachconformal,rubin2025conformal,rubin-toles2025conformal,cherian2024large}.
However, most of these algorithms operate through post-hoc filtering and require external LLM-based validation for scoring intermediate steps -- rendering them unsuitable for actively terminating generation.

In this work, we jointly pursue an effective and calibrated decision rule to determine when a language model can stop ``thinking.''
To do so, we introduce the notion of a reasoning tree, where at each step of sampling, a language model either adds a new leaf, walks along the tree, or backtracks to a previous step.
Notably, identifying when the thoughts have converged is equivalent to detecting when this reasoning tree stops growing.
Inspired by this concept, we approach early stopping as multiple hypothesis testing problem.
At each generation step, we test whether the current tree is expected to change, based on the predictions of lightweight probes over the language model's hidden representations.
Our algorithm is based on the Learn then Test framework~\citep{angelopoulos2021learn}, which provides finite-sample, distribution-free guarantees for controlling the risk of our decisions.

We evaluate this strategy, \emph{\ours{}}, based on its ability to guide efficient reasoning, and whether its decisions are well-calibrated.
Our experiments consider two empirical settings: we may or may not have access to training and calibration data from the true test distribution.
In the first setting, we train variants of \ours{} using three reasoning lanugage models (DeepSeek-R1 distilled Qwen 32B and Llama 70B~\citep{guo2025deepseek,yang2024qwen2,grattafiori2024llama}, QwQ 32B~\citep{qwq32b}), evaluated on a helf-out split of s1K-1.1~\citep{muennighoff2025s1}.
Here, we are able to halve the number of thinking tokens across accuracy levels, with a \textbf{maximum reduction of 60\%}.
Then we evaluate Qwen 32B-based \ours{} on three test datasets: AIME 24, GPQA Diamond~\citep{rein2024gpqa}, MATH-500~\citep{lightman2023let}).
Though these datasets vary in format and difficult, \ours{} is still able to reach \textbf{up to a 20\% reduction in thinking tokens}, and in the worst case, is \textbf{always as efficient as naive budget constraints}.
In summary, this work has three main contributions.
\begin{enumerate}
\item We interpret LLM reasoning through the lens of an abstract reasoning tree, where the problem of early exiting is equivalent to identifying when this tree stops growing.
\item This view allows us to calibrate the decision rule for actively terminating generation.
\item Based on multiple language models and reasoning benchmarks, we provide empirical evidence that \ours{} is effective for efficient test-time scaling.
\end{enumerate}

\section{Background}
\label{section:background}

\subsection{Test-time scaling}

% \paragraph{Scaling inference-time compute}

% There has been a recent shift for scaling language models from the pretraining phase to 

% It has been shown that the difficulty of the prompt used greatly influenced the effectiveness in scaling test-time compute, and that ``compute optimal'' test-time scaling is more effective than scaling model parameters~\citep{snell2024scaling}

\paragraph{Efficient inference}

Since current reasoning models are post-trained through reinforcement learning~\citep{guo2025deepseek}, a number of works address the overthinking problem~\citep{sui2025stop} as part of the reinforcement learning process~\citep{han2024token,arora2025training,hou2025thinkprune}.
Other works focus on the inference-time problem of predicting when a language model should stop generating~\citep{yang2025dynamic,zhang2025reasoning,ma2025reasoning}.
This papers falls under the latter category, which on a whole, is compatible with methods that reduce a language model's verbosity during post-training.
Finally, another option is to achieve efficiency in terms of model architecture.
Some works dynamically adapt compute cost~\citep{lei2021attention,leviathan2023fast}, while others employ only a subset of all modules during sampling~\citep{kim2021length,liu2022anytime,schuster2022confident}.
While these strategies operate over the Transformer stack, rather than the generation sequence length, many high-level ideas are broadly applicable to early exiting in our situation.

\paragraph{Self-consistency}

Self-consistency has been widely used to provide a self-supervised form of confidence during the sampling process~\citep{wang2022self}.
These methods aim to improve the quality of generated samples, often in situations where multiple samples may be sequentially generated~\citep{mitchell2022enhancing,madaan2023self,shi2023large,weng2023large,guo2025temporal}.
Consistency can also provide feedback for reasoning-focused reinforcement learning~\citep{wang2024math}.
Several recent works have observed that confidence scores can be probed and calibrated from internal representations, to prioritize reasoning trajectories for subsequent runs~\citep{liescape,huang2025efficient,xie2024calibrating} or for early exiting, similar to this work~\citep{zhang2025reasoning}.
Our key departure is that we calibrate the \emph{decision rule} to terminate generation, rather than the probabilistic outputs of a probe.
This reflects the online setting, where a probe is used to actively guide generation, rather than to filter trajectories post-hoc.

\subsection{Conformal prediction and risk control}
\label{sec:uncertainty}

Conformal prediction quantifies the uncertainty in machine learning models by generating set-valued predictions~\citep{shafer2008tutorial,angelopoulos2021gentle}.
These methods are distribution-free and valid under finite samples, which makes them particularly attractive in real-world applications.
Specifically, for an input $x$, a candidate output space $\mathcal{Y}$, and a predetermined error level $\epsilon$, conformal prediction tests each potential outcome $y \in \mathcal{Y}$ by evaluating the null hypothesis: ``output $y$ corresponds to input $x$.''
The final prediction set consists of the outputs $y$ for which this null hypothesis fails to be rejected, where the test statistic is known as a nonconformity score.
Split conformal prediction leverages a separate training set to learn this nonconformity score~\citep{vovk2005algorithmic,papadopoulos2008inductive}.
The true outcome is included with probability at least $1-\epsilon$, with guarantees that are typically marginal over draws of the test set and an exchangeable calibration set.

In the context of language modeling, conformal prediction has been adapted to calibrate the factuality~\citep{tatsu2024factuality,cherian2024large}, reasoning consistency~\citep{rubin2025conformal}, and quality of generations~\citep{quachconformal,semanticdensity}.
Here, $x$ may represent an input sequence of text, while $y$ may be a language model output.
Of these works, \citet{rubin2025conformal} also introduces the idea of reasoning as coherency over a graph structure, based on logical deducibility.
However, this and other methods are primarily designed for post-processing text that has already been generated, and they rely on external language models as scoring functions.
As a result, these approaches are not calibrated to be used as decision rules for iterative testing, and the latency required to compute nonconformity scores renders them unsuitable for early exiting.

More recently, the Learn then Test (LTT) framework~\citep{angelopoulos2021learn} extends the ideas in conformal prediction to control the risk of arbitrary loss functions, with guarantees over draws of the calibration set.
One application of LTT is to convert model outputs into a calibrated decision rule, by viewing hyperparameter selection (e.g. discretization thresholds) as multiple hypothesis testing.
Our method and several works in early exiting are built atop the LTT framework.
\citet{quachconformal} calibrates a language model's \emph{sampling} of output sets, similar to this work.
Their goal is to generate sufficient outputs $y$ until certain admissibility criteria have been fulfilled, e.g. correctness and diversity of information.
However, the sampling process in \citet{quachconformal} is interactive, in the sense that each step requires an external verifier, and text may be added or removed at any point.
As a result, this strategy is unsuitable for providing online decisions about when to stop.
\citet{schuster2022confident} also leverages LTT to calibrate a stopping rule to exit from a Transformer stack.
Their method operates on individual tokens, similar in spirit to applications like speculative decoding~\citep{leviathan2023fast}.
Our focus is on large, coherent thoughts for reasoning, where token-level uncertainties are less informative.

% More broadly, estimating and calibrating the uncertainty of language models has been well-studied in recent years.
% Methods range from 
% \cite{ji2025calibrating}``verbal uncertainty'' can be detected via a linear feature of the latent representation space in LLM
% \blue{hmmm}

% \cite{cherian2024large} extends traditional conformal methods by incorporating conditional validity adjustments and improved nonconformity measures, resulting in tighter and more context-sensitive confidence intervals for LLM-generated content. \cite{rubin2025conformal} leverages conformal prediction to ensure both the correctness and logical consistency of intermediate reasoning steps via prunning subgraphs, preserving coherent reasoning chains. 
% Finally, \cite{semanticdensity} proposes a method to quantify uncertainty by measuring semantic confidence directly in embedding spaces, enabling meaningful uncertainty assessments without explicit model logits.

\iffalse
conformal language modeling~\citep{quachconformal}

Language Models with Conformal Factuality Guarantees~\citep{tatsu2024factuality}

enhanced CP for LLM validity~\citep{cherian2024large}

Coherent factuality (pruning subgraph)~\citep{rubin2025conformal}

Semantic density~\citep{semanticdensity}

Graph-based metrics for LLM UQ~\citep{jiang2024graph}
\fi

\section{Thought Calibration}
\label{section:methods}

\begin{figure*}[t]
\centering
\includegraphics[width=\linewidth]{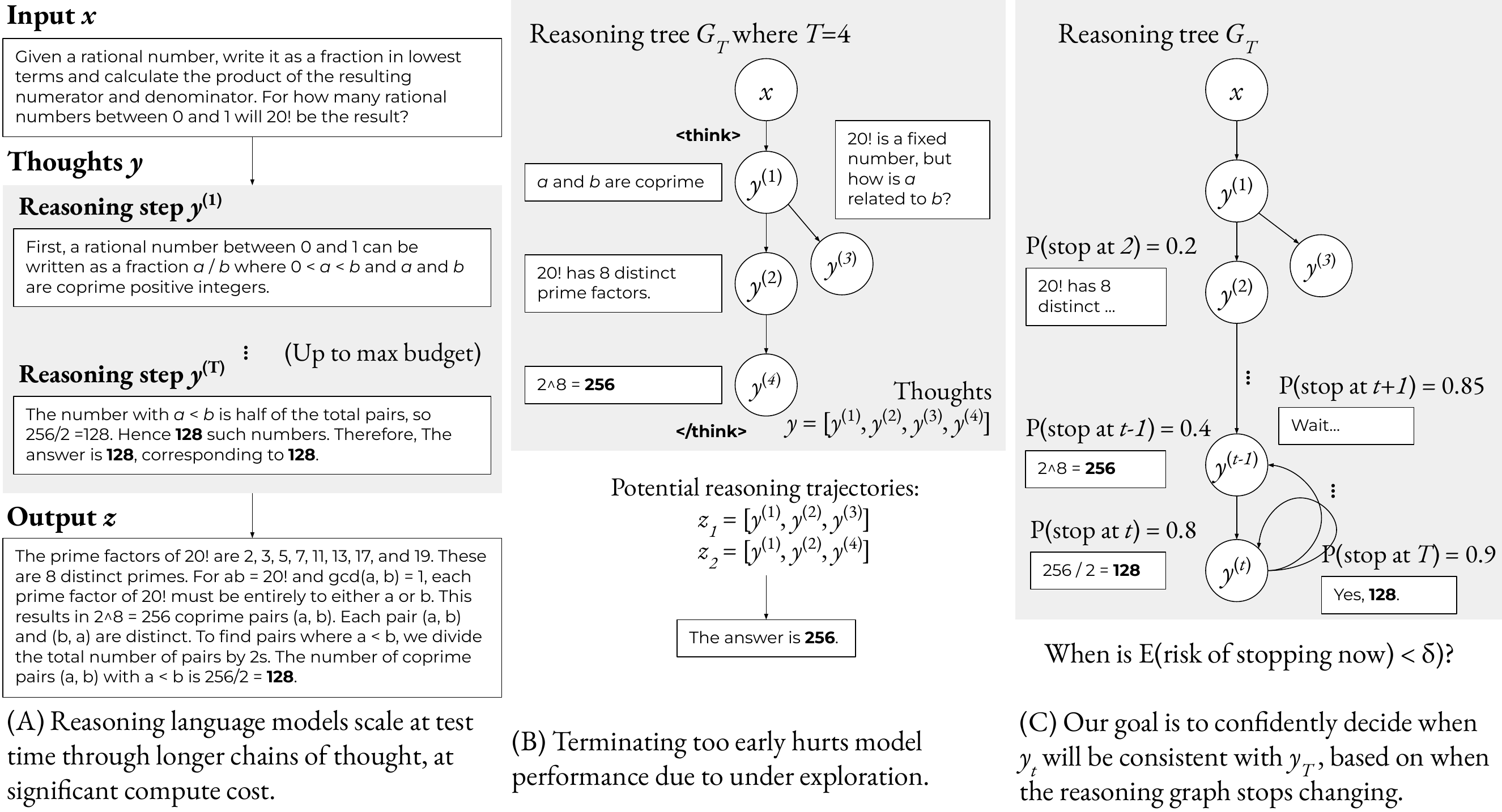}
\caption{Overview of the problem and our goal. Illustrated example based on s1K-1.1~\citep{muennighoff2025s1}.
}
\vspace{-0.2in}
\label{fig:overview}
\end{figure*}

Given an input $x\in\mathcal{X}$, a reasoning language model generates a series of thoughts $y\in\mathcal{Y}$, before synthesizing the final output $z\in\mathcal{Z}$.
For example, $x$ may represent a math question; $y$ is a sequence of reasoning steps; and $z$ is the model's attempt at solving the question (Figure~\ref{fig:overview}A).
Manipulating the budget allocated to generating $y$ directly impacts the quality of $z$~\citep{muennighoff2025s1}, but as the length of $y$ increases, so too does the cost of inference.
Our goal is to identify the point at which growing $y$ no longer improves $z$.

To formalize these ideas, we introduce the notion of an abstract \emph{reasoning graph} $G$, where nodes represent thoughts and directed edges represent entailment relationships~\citep{maccartney2014natural}.
This graph is rooted at $x$, the input question.
Nodes can be \emph{serialized} into textual descriptions, and different paraphrases of the same idea represent a single node.
Where it is clear, we refer to the abstract node and its textual representation interchangeably.

\begin{definition}
A \emph{reasoning trajectory} $z$ is a root-to-leaf walk in the reasoning graph $G$.
\end{definition}

An arbitrary $z$ need not be ``complete'' or ``correct'' with respect to the original question $x$. We use $z^*$ to denote a walk that starts at $x$ and ends at the right answer, which we assume to be incontrovertible.
$G$ uniquely determines the set of all root-to-leaf walks $\{ z \}$, and thus, whether  a language model has any chance of being correct in its final attempt.

\begin{definition}
A set of \emph{thoughts} $y$ is a walk, rooted at $x$, on the augmented graph $G'$ in which every node is connected to each of its ancestors.
\end{definition}

At each stage of sampling, a large language model either adds a leaf to $G$ (novel thought), or takes one step in $G'$ (backtracking or redundant generation).
Let $G_t$ be the reasoning graph at time $t$.
If a language model terminates thinking at this point, it is expected to answer correctly if there exists a path in $G_t$ that yields $z^*$.
Thus, it would be ideal we could calibrate the language model such that with high probability,
\begin{equation}
\label{eq:graph-answer}
\mathbb{P}\left(
 \mathbb{E} \left[
  \mathbbm{1}[
  z^* \not\in G_t
  ] \le \delta
 \right]
\right) \ge 1-\epsilon
\end{equation}
for some risk tolerance $\delta$ and error level $\epsilon\in(0,1)$.
In principle, a language model could enumerate the space of graphs in a combinatorial search.
However, it is far from guaranteed that this graph can be tractably found.
Instead, we focus on the consistency between reasoning graphs.
\begin{definition}
Thoughts $y$ and $y'$ are \emph{consistent} if they can be represented by the reasoning graph $G$.
\end{definition}

In particular, if a language model repeatedly revisits a step to arrive at the same conclusion, or traverses the same ideas in a different order, the resultant graph does not change (Figure~\ref{fig:overview}C).
Let $y_t := [y^{(i)} \dots y^{(t)}]$ and $G_t$ be the in-progress thoughts and reasoning graph after $t$ steps, and let $T$ be the maximum inference budget (token or model limit).
Instead of enforcing that $G_t$ contains $z^*$, it is more reasonable to guarantee that
\begin{equation}
\label{eq:graph-consistent}
\mathbb{P}\left(
 \mathbb{E} \left[
  \mathbbm{1}[
  G_t \ne G_T
  ] \le \delta
 \right]
\right) \ge 1-\epsilon.
\end{equation}
Due to the sequential nature of generation, $G_t$ is always a (not necessarily strict) subset of $G_T$.

We now describe how we calibrate the decision rule for terminating language model generation (Section~\ref{sec:ltt}), and then introduces three strategies for practically estimating the quantities described by Equations~\ref{eq:graph-answer} and \ref{eq:graph-consistent} (Section~\ref{sec:probes}).

\subsection{Calibrating the stopping rule}
\label{sec:ltt}

Suppose we have a calibration dataset $\mathcal{D}_\text{cal}$, which contains exchangeable points $\{ (x_i, y_i) \}_{i=1}^n$.
Given a new example $x$, let $y_t$ denote the language model's thoughts after $t$ sampling steps, and let $y_T$ denote the maximum set of thoughts.
Our goal is to find the smallest $t$ that fulfills Equations~\ref{eq:graph-answer} or \ref{eq:graph-consistent}, based on the distribution of $\mathcal{D}_\text{cal}$.
During the sampling process, however, we do not know $z^*$ or $G_T$, so we must estimate the quantities inside the expectation using a surrogate function $f$.
Here, $\mathcal{D}_\text{cal}$ serves to calibrate $f$ such that 
\begin{align}
\label{eq:realistic-answer}
\mathbb{P}\left(
 \mathbb{E} \left[
  R(y_t) \le \delta
  \mid \mathcal{D}_\text{cal}
 \right] 
\right) & \ge 1-\epsilon
\end{align}
where $R$ is a bounded risk function associated with $f$.
For example, $f$ may be a linear probe on the hidden representations of thought steps $y^{(i)}$, and its output may be a binary prediction.
A potential decision rule could take the form of a threshold $\lambda$, where if $f(y^{(t)}) \ge \lambda$, we terminate thinking.

% In classic conformal prediction, one might attempt to calibrate $f$'s uncertainty, yielding prediction sets $\subseteq\{0, 1\}$.
% However, this procedure does not directly yield a decision rule.
% Instead, when the classifier is queried iteratively about $y_t$, one requires a \emph{threshold} for 

Similar to \citet{schuster2022confident} and \citet{quachconformal}, we follow the Learn then Test framework to select a valid set of $\lambda$s that provide our desired guarantees~\citep{angelopoulos2021learn}.
On a high level, hyperparameter selection is viewed as a multiple hypothesis testing problem.
Let $\Lambda$ be a finite set of configurations, where each $\lambda_j\in\Lambda$ is associated with the null hypothesis,
\begin{equation}
H_j : \mathbb{E} [R(y_t) > \delta].
\end{equation}
The set of valid $\Lambda_\text{valid} \subseteq \Lambda$ is the set of $\lambda_j$ for which we \emph{fail to reject} $H_j$.
In particular, selecting the earliest stopping time is equivalent to identifying the smallest $\lambda\in\Lambda_\text{valid}$.

\begin{theorem}[Adapted from theorem 1 in~\citep{angelopoulos2021learn}]
\label{thm:ltt}
Suppose $p_j$ is super-uniform under $H_j$ for all $j$.
Let $\mathcal{A}$ be a family-wise error rate (FWER) controlling algorithm at level $\epsilon$.
Then $\Lambda_\text{valid} = \mathcal{A}(p_1,\dots,p_m)$ satisfies Equation~\ref{eq:realistic-answer}.
\end{theorem}

Theorem~\ref{thm:ltt} specifies that any FWER-controlling algorithm $\mathcal{A}$ can be used with an appropriate p-value to identify $\Lambda_\text{valid}$.
While \citet{angelopoulos2021learn} proposes several algorithms to search over $\Lambda$, we follow the fixed sequence testing method, since in principle, our risks are expected to be monotonic ($G_t\subseteq G_T$).

Specifically, let $\Lambda = \{\lambda_1, \dots, \lambda_m\}$ be a descending grid of parameters. Intuitively, larger $\lambda$ correspond to more permissive thresholds, e.g. allowing a language model to generate for longer.
\begin{enumerate}
\item For each $j$, we compute a valid p-value $p_j$, e.g. the binomial tail bound p-value, following~\citep{quachconformal}:
\begin{equation}
\label{eq:pval}
p_\lambda^{BT} :=
\mathbb{P}(
 \text{Binom}(n, \epsilon) \le
 n \hat R_n(\lambda)
).
\end{equation}
\item If $p_j \le \epsilon$, we reject $H_j$ and continue. Otherwise, we return $\lambda_{j-1}$ as the smallest valid threshold for error rate $\epsilon$.
\end{enumerate}
This process yields the binarization threshold for $f$, where we stop generating when $f(y_t) \ge \lambda_{j-1}$.

\subsection{Estimating empirical risk}
\label{sec:probes}

On a high level, the surrogate function $f$ should reflect the consistency of $y_t$ with expected future generations.
Ideally, we would be able to access the graphical structure of $G_t$, as any repetitions or redundant walks in $y_t$ would be immediately evident.
However, since autoregressive language models generate left-to-right, without explicitly conforming to any higher-level structure, we cannot operate directly over $G$.
Instead, we introduce three approaches for designing $f$ in practice.

We first briefly consider the simple case suggested by Equation~\ref{eq:graph-answer}: if we terminate thinking now, is the language model able to answer correctly?
That is, we could define
\begin{align*}
\label{eq:p_correct}
&f_\text{correct}(y_t) := ~\mathbb{P}(\text{\small LLM is correct based on } y_t) \numberthis
\\
&R_\text{correct}(y_t) :=
\mathbbm{1}\{\text{\small LLM is correct} 
\} \cdot (1 - f_\text{correct}(y_t))
\\
&~+\mathbbm{1}\{\text{\small LLM is wrong} 
\} \cdot f_\text{correct}(y_t).
\numberthis
\end{align*}
However, there are several drawbacks of this implementation.
By construction, the calibration dataset only contain questions that can eventually be answered, which is not true in general.
Though the space of graphs is countable, it is unlikely that a language model can efficiently explore the entire space.
In other words, the language model may realistically \emph{never} answer correctly.
Thus, setting $\lambda=1$ is \emph{not} guaranteed to be risk controlling.
With this definition of $R_\text{correct}(y)$, calibrating based on correctness also requires supervised labels.
While this is not an issue on standard benchmarks, it is harder to obtain labels (user feedback) in practice.

To address these challenges, we introduce two additional strategies for estimating graph consistency.
First, a language model's final attempt $z$ can be viewed as a distillation of its overall reasoning structure.
Thus, we compare the language model's attempt $z_t$ after $t$ steps, to the eventual attempt $z_T$ at the maximum reasoning budget.
This yields
\begin{align*}
\label{eq:p_consistent}
&f_\text{consistent}(y_t) := \mathbb{P}(z_t \text{\small~is the same as } z_T) \numberthis
\\
&R_\text{consistent}(y_t) := \mathbbm{1}\{\text{\small consistent} 
\} \cdot (1 - f_\text{consistent}(y_t))
\\
& +\mathbbm{1}\{\text{\small inconsistent} 
\} \cdot f_\text{consistent}(y_t)
\numberthis
\label{eq:r_consistent}
\end{align*}
These values can be determined even for intractable problems, as long as the extended reasoning produces no new insights, and does not require labels of correctness.

Finally, any particular $z$ only represents a single walk through $G$. Due to stochasticity, two differing attempts could be sampled from the same graph, which is no longer changing.
Towards this end, we observed that language models often reiterate redundant information, after having reached the correct answer or the extent of its abilities.
Probing for novelty should suffice to capture this phenomena.
In practice, however, we found that the following formulation was easier for our verifier to implement, as checking for novelty involves long context reasoning over all previous thoughts, which can be challenging~\citep{wang2024leave}.
% Thus we can model convergence in $G$ based on the lack of new leaves,
\begin{align*}
\label{eq:p_boring}
&f_\text{novel leaf}(y_t) :=
~\mathbb{P}(
y^{(t)} \text{\small is leaf}
) \cdot (1 - \mathbb{P}(
y^{(t)} \text{\small is novel}
))  \numberthis
\\
&R_\text{novel leaf}(y_t) := \mathbbm{1}\{\text{\small LLM inconsistent} 
\} 
\cdot f_\text{novel leaf}(y_t)
\\
&~+\mathbbm{1}\{\text{\small LLM consistent} 
\} \cdot (1 - f_\text{novel leaf}(y_t)).\numberthis \label{eq:r_boring}
\end{align*}
We reuse the labels for consistency due to ease of verification compared to novelty.

\subsection{Implementation details}
\label{sec:implementation}

% There are several gaps between the theory and implementation, which must be filled through design choices.
To separate a reasoning trajectory $y$ into individual steps $\{ y^{(i)} \}$, we use sections delimited by \texttt{\textbackslash n\textbackslash n}, which also contain either \texttt{wait} or \texttt{but}.
We observed that individual tokens representations can vary significantly.
Thus, each step uses the mean last-layer representation of its tokens, followed by dimensionality reduction via PCA to $d=256$.
% These sequences are concatenated back into partial trajectories for varying $t$.

To estimate each of quantities in \Cref{eq:p_correct,eq:p_consistent,eq:p_boring,eq:r_boring}, we train linear probes on these step-level representations.
The final probabilities are averaged over a window of 10 steps for smoothness, before calibration.
For evaluation, we use a grid of $\epsilon$ ranging from 0.05 to 0.5, with precise thresholds selected to roughly match the token range of baselines.
During development, we experimented with more complex architectures, e.g. Transformer to predict leaves as a sequence labeling task (Appendix~\ref{sec:other-probes}).
However, to avoid overfitting on our limited training set, we chose to focus on simple and efficient linear probes.
% However, due to our limited training set and based on calibration set accuracy, no variant significantly outperformed the linear probe to warrant the additional computational cost.
Concurrent work~\citep{zhang2025reasoning} also finds that model confidence can often be extracted linearly.
In our experiments, we use three reasoning models: DeepSeek-R1 distilled Qwen 2.5 32B and Llama 3.3 70B~\citep{guo2025deepseek,grattafiori2024llama,yang2024qwen2}, and QwQ 32B~\citep{qwq32b}.

The ground truth labels for these probes are obtained by prompting a separate language model (Qwen 3 32B).
\textbf{Correct}: We truncate thinking trajectories to desired lengths, append the \texttt{<\textbackslash think>} token, and prompt the language model for the final answer, which is compared to the ground truth~\citep{muennighoff2025s1}.
\textbf{Consistent:} The same outputs can be used to check whether $G_t$ is consistent with $G_T$, by comparing intermediate attempts $z_t$ to maximum budget attempt $z_T$.
\textbf{Leaf:} We annotate whether each step $y^{(i)}$ is a leaf in $G$ by asking a separate language model to identify whether it makes an attempt to answer the original question $x$, regardless of correctness.
\textbf{Novel:} We provide a separate language model with all previous thoughts $y^{(1)}\dots y^{(i-1)}$ and ask whether the new step $y^{(i)}$ provides additional information.
% Since full sets of thoughts can extensive, this task was the most difficult for our 32B verifier.
All prompts can be found in Appendix~\ref{sec:prompts}
and were run on 4 A6000 GPUs using vLLM~\citep{kwon2023efficient} and lmdeploy~\citep{lmdeploy}.

We evaluate the correctness of all final attempts using the GPT 4.1 API, between April 15, 2025 and May 15, 2025.
For datasets that have no ambiguity (multiple choice, numeric answers), we trimmed the final attempts to 200 characters, to prevent the LLM from ``cheating'' by using additional thinking budget after the \texttt{</think>} token.

\section{Experiments}
\label{section:experiments}

\begin{figure*}[t]
\centering
\includegraphics[width=\linewidth]{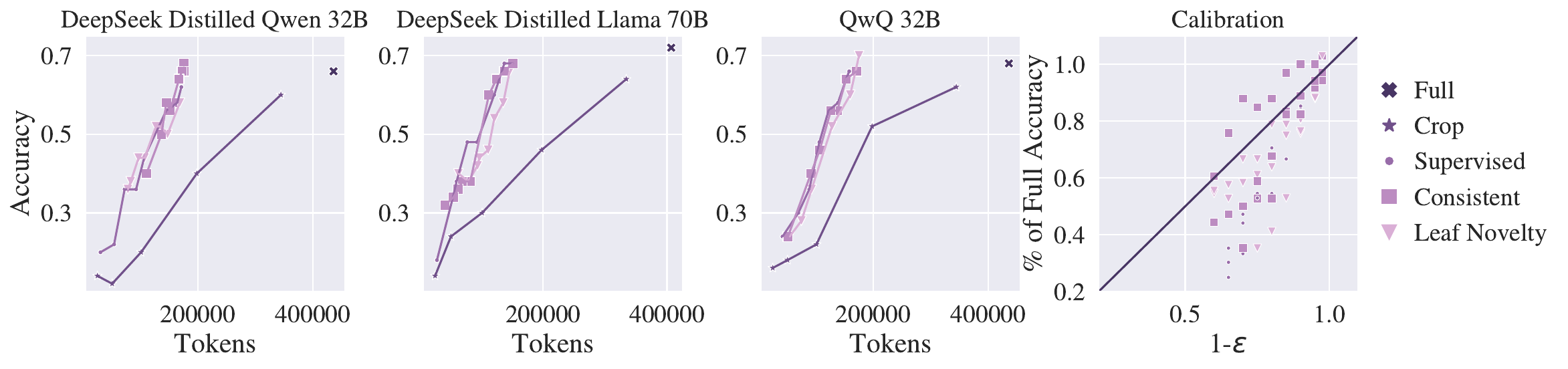}
\vspace{-0.3in}
\caption{On in-distribution data (held-out test split on s1K), variants of \ours{} achieve up to a 60\% reduction in thinking tokens while maintaining full performance.
Top right point: Complete DeepSeek-R1 thought trajectory from~\citep{muennighoff2025s1}.
Crop: Fix thinking budget at 512, 1024, 2048, 4096, and 8192 tokens.
Supervised: exit based on predicted likelihood of correctness.
Consistent, and Leaf Novelty: exit based on predicted consistency of answer or graph.
Supervised is over confident, since the test set contains unsolvable problems.
}
\vspace{-0.1in}
\label{fig:s1-trim}
\end{figure*}
\begin{figure*}[t]
\centering
\includegraphics[width=\linewidth]{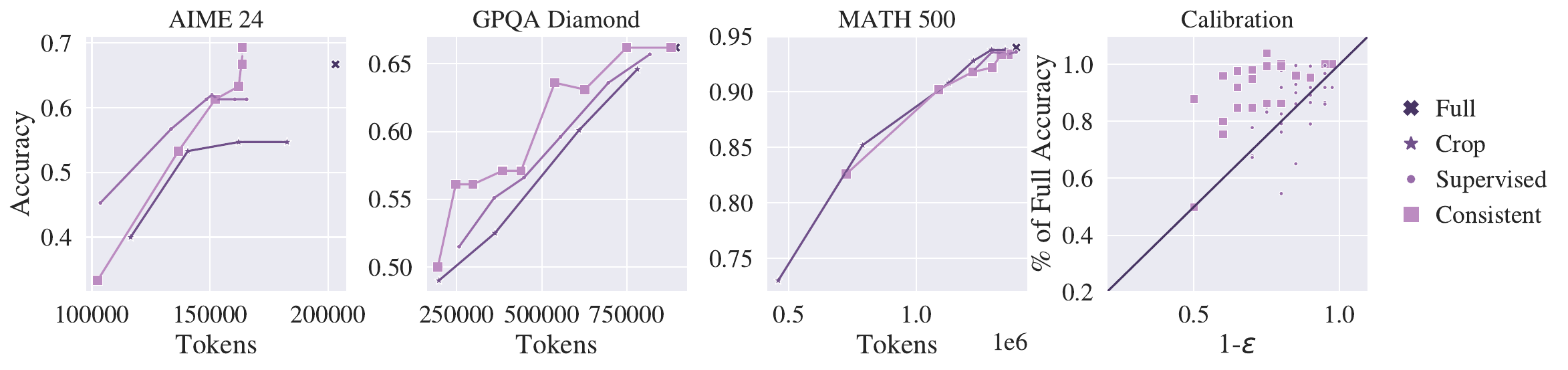}
\vspace{-0.3in}
\caption{We applied \ours{} probes for DeepSeek-distilled Qwen-2.5 32B on standard math and science benchmarks, which may be out-of-distribution compared to the training and calibration sets, drawn from s1K.
We achieve up to a 20\% reduction in thinking tokens.
While Consistent generally remains below the predetermined error rates, Supervised is overconfident (as expected).
}
\vspace{-0.2in}
\label{fig:test-set}
\end{figure*}

% \subsection{Datasets}
\subsection{Settings}

\paragraph{Datasets.}

Our experiments focus on efficient language model reasoning across tasks which vary in content, format, and difficulty.
In particular, we leverage the following datasets.

\textbf{s1K-1.1}~\citep{muennighoff2025s1} is a curated \emph{training set} for distilling reasoning abilities through data.
This dataset contains 1000 difficult math and science questions, along with thought trajectories generated by DeepSeek-R1~\citep{guo2025deepseek}.
As a proof of concept, we split the s1K-1.1 dataset into training, testing, and calibration (500, 50, 450, in dataset order).
We use the training set to develop our probes, which are calibrated on the calibration set and evaluated on the testing set.

We also consider three common reasoning benchmarks solely for testing.
\textbf{AIME-24} is the 2024 iteration of the American Invitational Mathematics Examination.\footnote{https://maa.org/maa-invitational-competitions/}
This dataset contains math questions whose answers take on integers between 0 and 999.
\textbf{GPQA Diamond}~\citep{rein2024gpqa} is a PhD-level math and science reasoning benchmark with multiple choice answers.
\textbf{MATH 500}~\citep{hendrycks2measuring,lightman2023let} is a curated subset of the MATH dataset, which competition math questions of various levels.
Note that while s1K-1.1 contains examples of both mathematical and scientific questions, the format and subsequent reasoning patterns may vary.
For example, while s1K-1.1 is open-ended, the various choices in GPQA must be compared.
Thus, we view these three datasets as ``out of distribution'' from s1K-1.1, which is itself diverse.

% \subsection{Models}
\paragraph{Models.}
We evaluate the three variants of \ours{}: the supervised probe for correctness (Equation~\ref{eq:p_correct}, \textbf{Supervised}); the consistency probe (Equation~\ref{eq:p_consistent}, \textbf{Consistent}); and the lack of novelty probe (Equation~\ref{eq:p_boring}, \textbf{Novel Leaf}).
To contextualize our experimental results, we also consider a naive budget-forcing baseline (\textbf{Crop}).
Specifically, we set a fixed token budget for thinking (ranging from 1024 to the full trajectory).
Once the language model reaches this budget, thinking is immediately terminated and the model is prompted for a final answer.
This reflects both the practical use case of setting a limit on maximum generation tokens, and the strategy employed by \citet{muennighoff2025s1}.
Finally, concurrent work has also observed that probes for correctness~\citep{zhang2025reasoning}
% or estimates of model confidence~\citep{yang2025dynamic} 
are effective for early exiting.
While this design may not be valid for risk control in practice (LLMs are not guaranteed to ever answer correctly), the \textbf{Supervised} baseline is similar to this work.

\subsection{In-distribution setting}

We start with the case where we have access to samples $x$ that are drawn from the same distribution as our eventual application.
For example, a model provider may possess typical examples of user data.
Our goals are to lower the overall test-time budget while maintaining accuracy, and to control any necessary drops in performance based on our predetermined error levels.
% Our goal is to lower the overall test-time budget while maintaining model performance, e.g. to reduce serving costs or customer-facing prices.
% At the same time, we would like to check whether our predetermined error level matches the empirical drop in performance.
In Figure~\ref{fig:s1-trim},
we observe that these probes are able to \textbf{reduce the number of thinking tokens by over half for all three models}, with minimal impact to overall performance.
With respect to calibration, the Supervised probe is quite poorly calibrated, especially at lower values of $\epsilon$.
All other probes are well calibrated at $\epsilon < 0.1$, though variance is higher outside of this range.
This may be due to distribution shift, resulting from the small test split (to maximize training and calibration data for subsequent evaluations).

\subsection{Generalization setting}

\begin{figure*}
\centering
\vspace{-5pt}
\includegraphics[width=\linewidth]{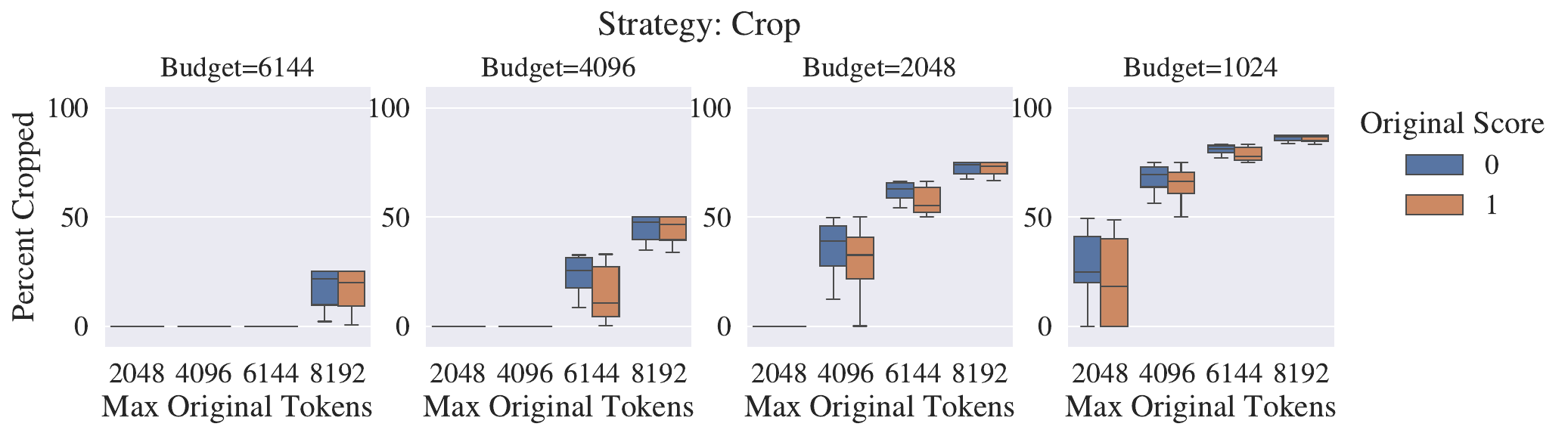}
\includegraphics[width=\linewidth]{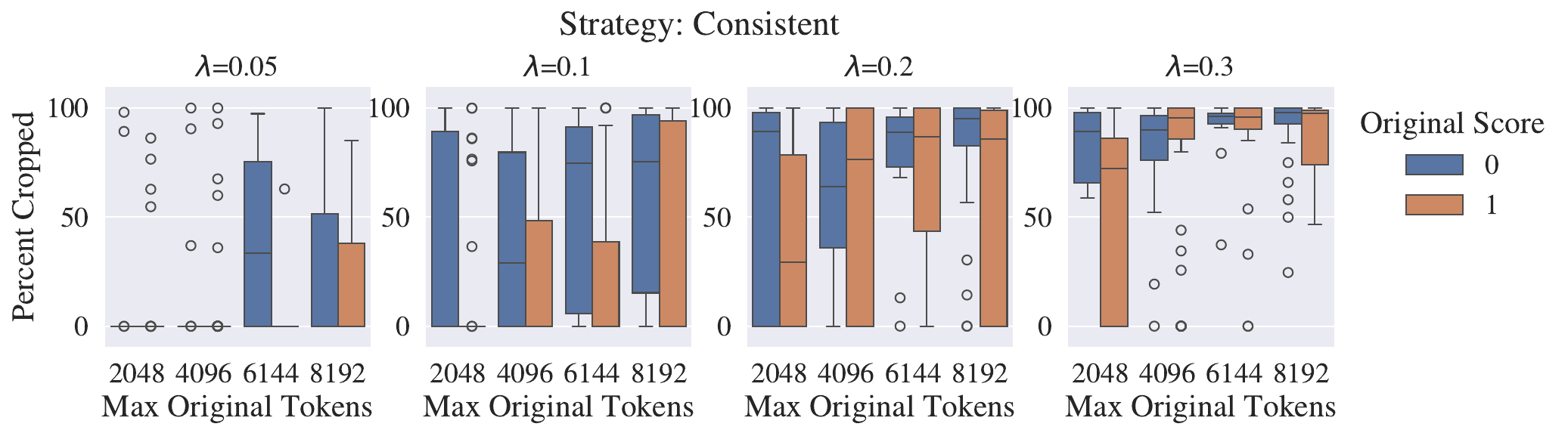}
\caption{
Proportion of prompt tokens removed, for different thresholds, stratified by full thought length and whether the original model was able to solve the problem.
Top: Naive max token thresholding.
Bottom: Consistency calibration, DeepSeek-R1 distilled Qwen 32B, over GPQA Diamond.
Cropping reduces token lengths uniformly, regardless of the input characteristics.
\ourscaps{} has a preference for first trimming longer thoughts and cases where the language model tries but fails to make progress.
}
\vspace{-5pt}
\label{fig:test-set}
\end{figure*}

Next, we consider the case in which the data we have is related, but \emph{not} drawn from the same distribution as our eventual application.
To emulate this setting, we apply the supervised and consistent Qwen 32B probes, developed on the s1K-1.1 dataset, to common reasoning benchmarks~\citep{rein2024gpqa,lightman2023let}.
Overall, we are able to improve (AIME 24, GPQA) or match (MATH 500) the efficiency of the budget forcing baseline -- even achieving slight \textbf{gains in performance on AIME 24}, perhaps by trimming distracting thoughts 
(Figure~\ref{fig:test-set}).
Notably, even though the Supervised probe had access to more information (ground truth answers), \textbf{the Consistent probe consistently generalizes better, both in terms of efficiency and calibration}.
Here, the Consistent probe fulfills the theoretical guarantees, while the Supervised probe remains over-confident.

\begin{figure*}
\centering
% \footnotesize
\vspace{-5pt}
\renewcommand{\arraystretch}{1.8}
\small
\begin{tabular}{>{\raggedright\arraybackslash}p{12cm}
>
{\raggedright\arraybackslash}p{\dimexpr \linewidth-2\tabcolsep-12cm}
}
\cellcolor[rgb]{0.9205787206936183, 0.9670402900213644, 0.9299288912264513}{Okay, so I need to solve this problem:   We have an increasing sequence: 3, 15, 24, 48, ... which are positive multiples of 3 that are one less than a perfect square. I need to find the remainder when the 1994th term of this sequence is divided by 1000.}
& Step 1
(0.104)
\\
\cellcolor[rgb]{1.0, 1.0, 1.0}{Hmm. Let me parse this. The terms are multiples of 3, and each term is one less than a perfect square [...] $n$ must be either 1 or $2 \mod 3$. That means terms in the sequence correspond to $n$ where $n \equiv 1$ or $2 \mod 3$.}
&
Step 2
(0.067)
\\
\cellcolor[rgb]{0.7489608747491013, 0.8835901392092831, 0.7760543758861975}{But computing $29922 \mod 1000$ is equivalent to [...]  but since $2992=2000+992=2000+992$, but modulo $1000, 2992 \equiv 992 \mod 1000$, so $992^2 - 1 \mod1000$.}
&
Step 37
(0.342)
\\
\cellcolor[rgb]{0.5082193274323427, 0.728623796805286, 0.5525746818608226}{Wait, first compute $992^2 \mod1000$, then subtract 1, then take $\mod1000$ again. [...] subtract 1: $64 - 1 = \textbf{63}$.  Therefore, $9922 -1 \mod1000=\textbf{63}$. Therefore the remainder is \textbf{63}. So answer is \textbf{63}.}
&
Step 38
(0.717)
\\
\cellcolor[rgb]{0.5501852107235036, 0.7596905579845981, 0.5923471685044214}{But let me confirm because that seems straightforward. Wait: [...]}
&
Step 39
(0.646)
\\
\cellcolor[rgb]{0.65380101476573, 0.8285709484671111, 0.6889726352633602}{Wait: $n(k)= (3k)/2+1$ for even $k$. For even $k=1994$, [...] Then term = $29922 - 1$. Then $\mod 1000$ is $(29922 - 1) \mod1000$.}
&
Step 40
(0.479)
\\
\cellcolor[rgb]{0.35894078277961305, 0.5989873014869181, 0.4072490065051902}{But $2992 \mod 1000 = 992$, so $2992\equiv - 8 \mod1000$. Then $(-8)^2=64$, then $64-1=\textbf{63}$. Therefore mod 1000: \textbf{63}. [...] Then $(-8)^2=64$, then $64-1=\textbf{63}$. Therefore mod 1000: \textbf{63}.  Hence remainder is \textbf{63}.}
&
Step 41
(0.985)
\end{tabular}

\caption{DeepSeek-R1 distilled Llama 70B Consistency probe on s1K-1.1 example from our test split, where color intensity is proportional to $\mathbb{P}(\text{consistent})$.
The language model first reaches the correct answer in Step 38, backtracks with lower confidence, and returns to the answer in Step 41.}
\label{fig:highlight}
\end{figure*}

\subsection{Additional analysis}

Figure~\ref{fig:test-set} illustrates that \ours{} probes prioritizes the termination of problems which cannot be solved, even at full budget -- perhaps hinting that the language model may have been stuck in a cycle of reasoning, without novel progress.
Compared to the naive cropping strategy, \ours{}'s input-dependent decision also demonstrate significant variance in the amount of tokens across different problems.

We also examine a specific instance from our s1K-1.1 testing split in Figure~\ref{fig:highlight} (s1K is a distillation dataset, so this diagram does not leak real test examples).
The language model reaches the correct answer after 38 steps (out of 48 steps).
As the model backtracks, the predicted consistency (with the expected final answer) drops; and as the model returns to the answer, confidence increases, higher than before.
This reaffirms that self-consistency is indeed a powerful indication of correctness, both distilled into a predictive model, and over the course of sampling.

\section{Limitations}

There are several limitations of our work.
Since our method is built atop the Learn then Test framework~\citep{angelopoulos2021learn}, our theoretical guarantees are only valid over draws of the calibration set.
In practice, this means that the calibration data must be sufficiently similar to the actual application.
Furthermore, due to our small training and calibration datasets, we implement our framework primarily through linear probes.
In Appendix~\ref{sec:other-probes}, we found that more complex architectures may lead to slightly better performance in some cases, and the gap is expected to be larger if more training data can be gathered.
We leave further investigations regarding the probe architecture to future work.
Finally, this paper only addresses the problem of exiting early from reasoning.
The broader question of how to calibrate the \emph{steering} of reasoning models remains unanswered, and is an interesting area for further research.

% \section{Discussion}
% \label{section:results}

\newpage

\bibliography{references}

\newpage
\newpage
\appendix
\onecolumn

\section{Prompts}
\label{sec:prompts}

The following prompt was used to force the model (DeepSeek-R1 distilled Qwen 32B and Llama 70B, QwQ 32B) to produce an answer after a fixed number of thinking steps.
Following the recommendation of \citet{guo2025deepseek} and \citet{qwq32b}, we do not include a system prompt.
We apply the chat template to user prompt before concatenating the ``in-progress'' thoughts.
Adapted from~\citep{muennighoff2025s1}.

\begin{prompt}
<bos><User>

\fstring{question}

Please reason step by step, and put your final answer within \textbackslash\textbackslash boxed\{\{\}\}.

<Assistant>

<think>

\fstring{thoughts}

</think>

Final Answer:
\end{prompt}

The following prompt was used to obtain labels for $\mathbb{P}(\text{correct})$ (Equation~\ref{eq:p_correct}) using Qwen 3 32B.
This prompt was also used to evaluate answers using GPT 4.1.
Adapted from~\citep{muennighoff2025s1}.

\begin{prompt}
You are an AI assistant for grading a science problem. The user will provide you with the question itself, the correct answer, and the student's attempt. Your job is to judge whether the attempt is correct by comparing it with the correct answer. If the correct answer is a number or choice, there should be no ambiguity, and you should directly compare the answer and the final result. If the attempt is incomplete, you should mark it as wrong. If the correct answer involves going through the entire reasoning process, you should judge the result based on whether the reasoning process is correct, compared to correct answer.

Do NOT try to solve the problem yourself. Only grade the attempt based on the correct answer.

The user will provide the attempt and the correct answer in the following format:

\# Problem

\fstring{problem}

\#\# Correct answer

\fstring{solution}

\#\# Student attempt

\fstring{attempt}

Explain your reasoning concisely, and end your response on a new line with only "Yes" or "No" (without quotes).
\end{prompt}

The following prompt was used to obtain labels for $\mathbb{P}(\text{consistent})$ (Equation~\ref{eq:p_consistent}) using Qwen 3 32B.

\begin{prompt}
You are an AI assistant for grading a science problem. The user will provide you with the question itself and two student attempts. Your job is to judge whether the two students arrive at the same answer. If question asks for a single numerical answer, there should be no ambiguity, and you should directly compare the two answers. If the question asks for multiple parts, the two attempts are identical if only if all of the parts arrive at the same conclusion.

Do NOT try to solve the problem yourself. Only grade whether the two attempts are the same.

The user will provide the problem and two attempts in the following format:

\# Problem

\fstring{problem}

\#\# Attempt 1

\fstring{attempt1}

\#\# Attempt 2

\fstring{attempt2}

Explain your reasoning concisely, and end your response on a new line with only "Yes" or "No" (without quotes).
\end{prompt}

The following prompt was used to obtain labels for $\mathbb{P}(\text{leaf})$ (Equation~\ref{eq:p_boring}) using Qwen 3 32B.

\begin{prompt}
You are an AI assistant for parsing LLM outputs. The user will provide you with the question and an intermediate reasoning step. Your job is to judge whether the given step contains an attempt at a final answer.

Do NOT attempt to solve the problem yourself. It does not matter if the answer is correct. Only comment on whether an attempt has been made.

The user will provide the problem and reasoning steps in the following format:

\# Problem

\fstring{problem}

\# Reasoning step

\fstring{reasoning step}

Explain your reasoning, and end your response on a new line with only "Yes" or "No" indicating whether or the given step makes an attempt at providing the final answer.
\end{prompt}

The following prompt was used to obtain labels for $\mathbb{P}(\text{novel})$ (Equation~\ref{eq:p_boring}) using Qwen 3 32B.

\begin{prompt}
You are an AI assistant for assessing the quality of logical reasoning. The user will provide you with the question and an incomplete attempt, consisting of a series of reasoning steps. Your job is to judge whether current step appears to provide additional information, compared to the previous steps. If the current step is correct and novel, it is useful. If the current step is wrong or redundant, then it is not useful.

Do NOT try to solve the problem yourself. It does not matter if the attempt is not complete. Only comment on whether the current step is useful.

The user will provide the problem and reasoning steps in the following format:

\# Problem

\fstring{problem}

\# Reasoning

\#\# step 1

\fstring{reasoning step 1}

\#\# step 2

\fstring{reasoning step 2}

...

\#\# step k

\fstring{reasoning step k}

...

\#\# current step

\fstring{current reasoning step}

Explain your reasoning, and end your response on a new line with only "Yes" if the current step provides new information or "No" otherwise (without quotes).
\end{prompt}

\section{Implementation details}

\subsection{Design and implementation of model probes}
\label{sec:other-probes}

We tried several architectures, before deciding upon linear probes for simplicity and to avoid overfitting. The differences in performance are not always consistent and the generalization gap is quite large (Table~\ref{table:probe-results}).
Since our main focus is on calibration, and it requires significant compute to produce and evaluate scaling curves, we consider more exhaustive exploration of alternate architectures as future work.

\paragraph{MLP}
The input is a single representation $h^{(t)}$ corresponding to single reasoning step $y^{(t)}$, and the output is a binary label $\in\{0, 1\}$.
We train until AUC fails to improve for 10 epochs on 10\% of the training set (randomly sampled).
We report the best calibration set performance of the following hyperparameters.
We use the sklearn defaults otherwise~\citep{scikit-learn}.
\begin{itemize}
\item Layers: 1, 2
\item FFN dimension: 32, 64, 128
\end{itemize}

\paragraph{Transformer}
The input is a sequence of representations, $h^{(1)} \dots h^{(t)}$ corresponding to thoughts $y_t = y^{(1)} \dots y^{(t)}$.
The output is either a binary label $\in \{0, 1\}$ for $\mathbb{P}(\text{correct})$ and $\mathbb{P}(\text{consistent})$, or a sequence of labels $\in \{0, 1\}^t$ for $\mathbb{P}(\text{novel})$ and $\mathbb{P}(\text{leaf})$.
For the former, we treat the embeddings as a set (i.e. if \emph{any} representation is sufficient to answer correctly, or be consistent).
For the latter, we apply a left-to-right causal attention mask during training, and we use sinusoidal positional encodings to encode the index of each reasoning step.
We report the best calibration set performance of the following hyperparameters.
In contrast to the linear and MLP models, we find that the Transformer performs best if we \emph{do not} apply PCA and instead operate over the original model dimension.
\begin{itemize}
\item Layers: 1, 2
\item Model dimension: 16, 32, 64
\item FFN dimension: 64, 128
\item Number of heads: 4, 8
\item Epochs: 5, 10
\end{itemize}

\begin{table*}[ht]
\caption{Probe architecture performance on s1K-1.1 train and calibration splits. Metric: Binary AUROC.}
\label{table:probe-results}
\begin{center}
\begin{small}

\begin{tabular}{>{\raggedright\arraybackslash}p{2cm} l rr rr rr}
\toprule
&
& \multicolumn{2}{c}{Linear}
& \multicolumn{2}{c}{MLP}
& \multicolumn{2}{c}{Transformer}\\
\cmidrule(l{\tabcolsep}){3-4}
\cmidrule(l{\tabcolsep}){5-6}
\cmidrule(l{\tabcolsep}){7-8}
Model &
Quantity &
Train & Cal &
Train & Cal &
Train & Cal
\\
\midrule

\multirow{4}{2cm}{DeepSeek-R1 distilled Qwen 2.5 32B} &

$\mathbb{P}(\text{correct})$ &
0.936 & 0.788 &
0.990 & 0.779 &
0.994 & 0.760

\\

& $\mathbb{P}(\text{consistent})$ &
0.919 & 0.788 &
0.994 & 0.747 & 
0.991 & 0.773
\\

& $\mathbb{P}(\text{leaf})$ &
0.868 & 0.839 &
0.936 & 0.815 &
0.933 & 0.852

\\

& $\mathbb{P}(\text{novel})$ &
0.874 & 0.686 &

0.980 & 0.692 &

0.896 & 0.774

\\

\midrule

\multirow{4}{2cm}{DeepSeek-R1 distilled Llama 3.3 70B} &

$\mathbb{P}(\text{correct})$ &
0.937 & 0.765 &
0.987 & 0.746 &
0.991 & 0.803
\\

& $\mathbb{P}(\text{consistent})$ &
0.921 & 0.745 &
0.994 & 0.743 &
0.993 & 0.748
\\

& $\mathbb{P}(\text{leaf})$ &
0.864 & 0.819 &
0.970 & 0.802 &
0.923 & 0.848

\\

& $\mathbb{P}(\text{novel})$ &
0.872 & 0.686 &
0.981 & 0.702 &
0.915 & 0.774

\\

\midrule

\multirow{4}{2cm}{QwQ 32B} &

$\mathbb{P}(\text{correct})$ &
0.943 & 0.848 &
0.986 & 0.838 &
0.948 & 0.848
\\

& $\mathbb{P}(\text{consistent})$ &
0.950 & 0.699 &
0.988 & 0.704 &
0.939 & 0.756
\\

& $\mathbb{P}(\text{leaf})$ &
0.869 & 0.840 &
0.942 & 0.822 &
0.913 & 0.857
\\

& $\mathbb{P}(\text{novel})$ &
0.876 & 0.677 &
0.952 & 0.690 &
0.895  & 0.792

\\

\bottomrule
\end{tabular}
\end{small}
\end{center}
\vskip -0.2in
\end{table*}

\subsection{LLM experiments}

We ran DeepSeek-R1 distilled Qwen 2.5 32B and Llama 70B, and QwQ 32B using lmdeploy~\citep{lmdeploy} with recommended defaults for each model.
lmdeploy natively supports the saving of last layer representations, so it was used for almost all experiments.
We ran Qwen 3 32B using vLLM~\citep{kwon2023efficient} due to early support.
Due to computational constraints, we report the mean over a single run.

We downloaded all model weights from \texttt{transformers} between April 1, 2025 and May 1, 2025.

\section{Additional analysis}

\begin{figure*}[h]
\centering
\includegraphics[width=\linewidth]{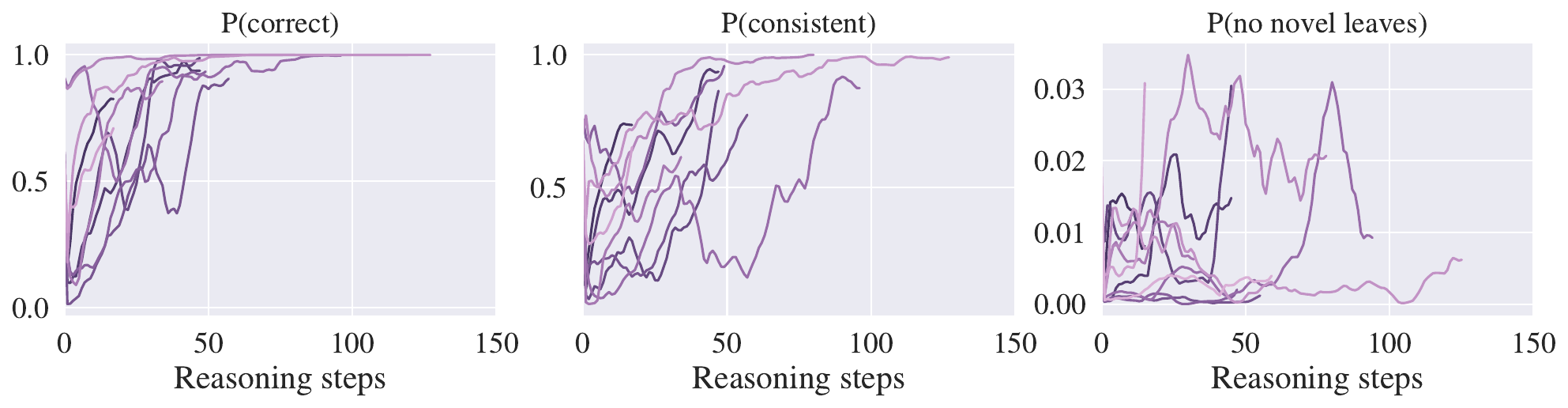}
\caption{Likelihoods of \ours{} probes over s1K-1.1 test set (10 examples).
The ``No Leaf'' variant is the least monotonic.
This could potentially indicate that after reaching the answer, the language model explores new knowledge that is irrelevant to the task.}
\label{fig:p_over_time}
\end{figure*}

Figure~\ref{fig:p_over_time} illustrates the early exit probabilities for each of the three probes.
The supervised (``correct'') probe reaches high exit probabilities the fastest, but it is also the most overconfident (Figure~\ref{fig:s1-trim}D).

\end{document}